# Data-driven Flood Emulation: Speeding up Urban Flood Predictions by Deep Convolutional Neural Networks

Zifeng Guo[1], João P. Leitão[2], Nuno E. Simões[3], and Vahid Moosavi[1]

[1]Swiss Federal Institute of Technology Zurich (ETHZ), Switzerland.

[2]Swiss Federal Institute of Aquatic Science and Technology (Eawag), Switzerland.

[3]University of Coimbra, Portugal.

guo@arch.ethz.ch
JoaoPaulo.Leitao@eawag.ch
nuno.simoes@dec.uc.pt
moosavi@arch.ethz.ch

## Abstract

Computational complexity has been the bottleneck for applying physically based simulations in large urban areas with high spatial resolution for efficient and systematic flooding analyses and risk assessments. To address this issue of long computational time, this paper proposes that the prediction of maximum water depth rasters can be considered an image-to-image translation problem where the results are generated from input elevation rasters using the information learned from data rather than by conducting simulations, which can significantly accelerate the prediction process. The proposed approach was implemented by a deep convolutional neural network trained on the flood simulation data of 18 designed hyetographs in three selected catchments. Multiple tests with both designed and real rainfall events were performed, and the results show that flood predictions by neural networks use only 0.5% of the time compared with that of physically based approaches, with promising accuracy and generalizability. The proposed neural network can also potentially be applied to different but relevant problems, including flood predictions for urban layout planning.

# 1 Introduction

The combination of the increased urbanization rate and the increased rainfall intensity due to climate change has posed great challenges for flood risk management (Plate, 2012), in which fast prediction methods are demanded for conducting systematic analyses and investigations of different scenarios. Flood occurrence likelihood can be estimated by rapid water depth predictions in different rainfall patterns, and rainfall spatial variation is also a driver for efficiently exploring different scenarios (Zheng et al., 2015). Furthermore, rapid flood predictors can be integrated with online weather forecast services in the future to inform citizens in advance of likely flood events so that precautionary measures can be taken.

The current bottleneck for rapid flood analyses is the long computational time required by physically based simulation models. This issue becomes extremely significant when running simulations on large areas or with high spatial resolution. According to Teng et al. (2017), physically based models are not suitable for areas larger than 1,000 $km^2$ when the spatial resolution is greater than 10 m. In urban areas, the feasible size for simulations is much smaller, as the required resolution increases to between 1 m and 5 m (Fewtrell et al., 2008 and Leitão et al., 2009). Although many efforts have been made to develop faster physically based (e.g., Bradbrook et al., 2004; Chen et al., 2007) and non-physically based (e.g., L'homme et al., 2008; Teng et al., 2015; Guidolin et al., 2016; Jamali et al., 2019) simulation models, state-of-the-art methods are still too slow for most urban-size applications, especially for those such as flood-driven optimization where iterative analyses are necessary.

In this paper, we propose that the prediction of maximum water depth can be regarded as an image-to-image translation problem in which the output water depth raster is generated directly from the input elevation raster using a convolutional neural network (CNN) (Fukushima, 1988). Compared with other types of neural networks, the CNN utilizes the information of adjacent pixels of the input image (raster) with much fewer trainable parameters and therefore is extremely suitable for solving image-based problems. The designed neural network was trained with a dataset of simulation results of 18 design hyetographs in three selected catchment areas, and validations on both design and real rainfall events were performed. The generalizability of the CNN makes the trained network useful for fast systematic flood investigations and analyses.

# 2 Previous Related Studies

## 2.1 Flood modelling

Numerical flood simulation models can be categorized as physically based and non-physically based approaches. The first category includes methods that typically require a set of differential equations to be (numerically) solved. They can be further categorized depending on the dimensionality of the flow representation. One-dimensional (1D) models simulate the flow along a centreline and are suitable for situations such as in a confined channel or in a pipe (e.g., Samuels, 1990). Two-dimensional (2D) models represent the flow as a two-dimensional field where the water depth is assumed to be shallow compared with the other spatial dimensions (e.g., Bradbrook et al., 2004; Chen et al., 2007; Bates et al., 2010). 2D models are considered sufficient for most of the applications except those such as dam breaking and tsunamis (e.g., Monaghan, 1994; Ye & McCorquodale, 1998), where vertical features are critical and three-dimensional (3D) models are demanded. Many efforts have been made to speed up physically based models, and a common strategy is to reduce the complexity of the differential equations, for example, by neglecting the inertial and advection terms of the momentum equation (e.g., Bradbrook et al., 2004; Chen et al., 2007) or decoupling the flow into orthogonal directions (e.g., Hunter et al., 2005; Bates et al., 2010). Hunter et al. (2007) provided a detailed theoretical discussion and review regarding models with reduced complexity.

The non-physically based approaches (e.g., L'homme et al., 2008; Jamali et al., 2019) predict water depths based on simplified hydraulic concepts. This category produces approximate predictions of the inundation results with much less computational cost compared with that of physically based models (Hunter et al., 2008, Néelz & Pender, 2010 & 2013). Therefore, they are suitable for applications for which some flow properties, such as velocity, are not required. Recently, cellular-automata flood models (e.g., Ghimire et al., 2013; Guidolin et al., 2016) have received considerable attention. They are a group of methods that model complex physical phenomena by transition rules rather than by solving a system of differential equations. The transition rules calculate the new state of a cell from the previous state of itself and its neighbours. The rules are applied on all raster cells in parallel, which benefits from parallel techniques such as GPU acceleration and therefore can significantly reduce the simulation time.

## 2.2 Neural networks and machine learning

Recent advances in artificial intelligence have shown that learning from data can be effective in applications such as classifications and object detections in which making predictions using domain knowledge may be difficult. Many methods have been developed to generalize from training data, among which artificial neural networks are one of the most prominent categories. Artificial neural networks are composed of computational units that combine the input and weight vectors and produce the output value via optional activation functions for thresholding. The different ways in which computational units connect are called layers, which can stack one after the other to build the entire neural network (Haykin, 1994). Early neural networks were limited in their ability to process data in their raw form, especially for high-dimensionality images, and thus require considerable domain expertise for data preprocessing and careful engineering for network design (LeCun et al., 2015). Recently, the backpropagation procedure (Werbos 1974) has shown that multiple-layer neural networks can be trained by a simple gradient decent method, making joint training of all the layers possible (Bengio et al., 2013).

The significant impact of neural networks and machine learning on complex tasks can be observed from the field of computer vision, in which tasks such as image classification (e.g., LeCun et al., 1998) can be difficult to perform based on explicit domain knowledge and rules. Recently, online data repositories such as ImageNet (Deng et al., 2009; Russakovsky et al., 2015) have boosted research with different approaches. Examples include AlexNet (Krizhevsky et al., 2012), VGG-Net (Simonyan & Zisserman, 2014) and GoogLeNet (Szegedy et al., 2015) for image classification tasks; the YOLO model (Redmon et al., 2016) for object detection; the FCN model (Long et al., 2015); the SegNet model (Badrinarayanan et al., 2017); and the pix2pix model (Isola et al., 2017) for semantic segmentation (to predict the label for each pixel of the input image). In particular, semantic segmentation has been very similar to flood prediction tasks, as both convert one image (raster) to another. Therefore, it would be very promising if relevant techniques could be migrated.

## 2.3 Data-driven flow and flood modelling

Using computer simulations to generate data has become a common alternative for flow-related research due to the lack of large and complete real datasets. For example, particle-based simulations were used to train a regression forest to predict the next state of liquid particles from the previous states for making frame-by-frame liquid animations (Jeong et al., 2015). The regression model acted as a surrogate for the time-consuming simulations and therefore was able to produce animations of millions of liquid particles with user interactions in real-time. Hennigh (2017) proposed a method to encode the previous states of the flow and predict the following states by a series of weight-shared convolutional autoencoders. A similar strategy was followed by Singh et al. (2017), who used a CNN for the inference of the flows of different airfoils represented by binary images. Raissi et al. (2019) used a feed forward neural network to explicitly approximate the Navier-Stokes equations from scattered data, and the results showed a high level of precision compared with simulation results. However, the time-related performance of the model was not reported. As a brief conclusion, these studies have shown promising results, but they cannot be directly applied to applications of flood prediction, as flood simulators are not particle-based models, and the main objective is to predict the water level regardless of the exact flow of water molecules.

Regarding flood-related research, Mustafa et al. (2018) proposed a single-layer fully connected neural network to predict the average water depths from the parameters for creating terrains. Similar work can be found by Feng et al. (2016), who used a regression model to predict the pedestrian comfort level based on the parameters for generating building layouts. However, it is difficult to directly extend these approaches to other applications because the inputs of the neural networks were the parameters of specific generative algorithms rather than the raw elevation data, which creates large differences in the input dimensionality and consequently the techniques to avoid the exponential explosion of the number of trainable parameters. Zaghloul (2017) proposed a method based on a self-organizing map (SOM) (Kohonen, 1990) to predict the wind velocities under uniform inlet flow around buildings from geometric features extracted by a ray-shooting algorithm. This approach was further investigated to predict the overland water depth (Leitão et al., 2018), and the accuracies of the results were promising. However, the ray-shooting method could be very inefficient for feature extraction when the resolution increases, and the postprocessing step of "smoothing" the discrete raw prediction due to the limited number of SOM cells may reduce the prediction accuracy. Therefore, we propose the use of a CNN to learn directly from the raw raster pixels.

## 3 Problem Statement

The amount of computational time required by physically based simulations has emphasized the need for developing fast flood prediction tools for urban flood risk management and urban planning. In particular, these applications are mainly concerned with the worst flooding cases rather than the dynamic process of the flow. Therefore, we neglect the flow dynamics and focus on the prediction of the maximum water flood level. This task can be further divided into more specific problems for different applications, for example, predicting the water depth 1) at the same coordinate resulting from different rainfall patterns, 2) of different places by the same rainfall pattern, and 3) considering both the terrain and rainfall variations. As a first step, we focus only on the maximum water depths of specific catchment areas under given hyetographs to highlight the generalizability of CNNs for flood prediction. Relevant works (e.g.,

Hennigh, 2017 and Singh et al., 2017) have shown that flow velocity can be included by adding extra image channels to the output, and the predictions between different image channels are relatively independent.

## 4 Proposed Approach

### 4.1 Framework

The proposed approach predicts the maximum water depth of one specific catchment from input hyetographs and elevation data. The key idea is that the flood prediction task can be regarded as an image-to-image translation problem where both the input and output are images (rasters) of different data. The pipeline of our approach can be described in three main steps. First, the terrain rasters are preprocessed and encoded with multiple image channels for different surface features. The rainfall hyetographs are sampled every 5 minutes for a one-hour period and represented as 12-dimensional vectors. Second, patch locations are randomly sampled from the catchment area, and the terrain and water depth data within the patch locations are extracted (called terrain and water patches). They are split into training and test sets according to corresponding hyetographs for training and validating the CNN model. The patch size equals the input size of the neural network. After the training, the flood predictions for new rainfall vectors are performed using new patch locations that are obtained based on an orthogonal grid rather than random sampling to minimize the number of patches needed to cover the entire catchment. Water depths are predicted for these patch locations using terrain data from the same locations and the new rainfall vector. Finally, the resulting water patches are assembled as the final output. The reasons for working with patches and not the entire catchment areas include the following: 1) patches may share similarities that help the machine learn and generalize; 2) this approach is an effective way to produce more training data from a limited number of simulation results; and 3) each catchment raster contains millions of cells, and running CNN on this size could be slow and memory consuming. Figure 1 shows the framework of our approach.

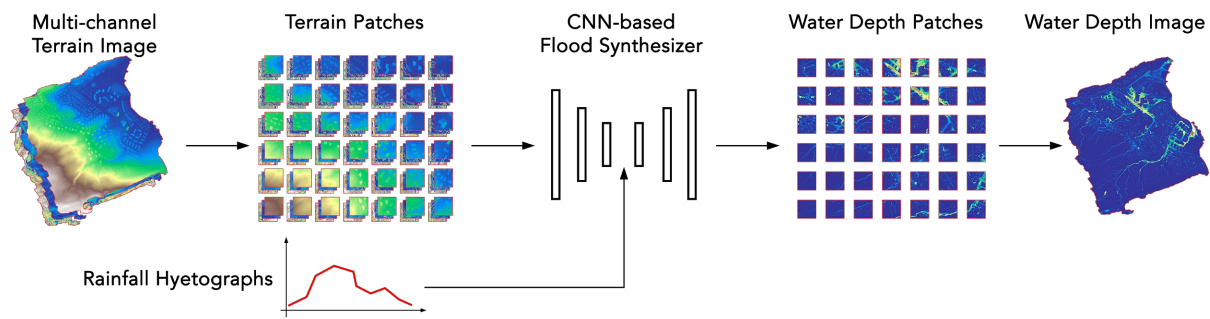

Figure 1. Data-driven flood emulation framework

### 4.2 Catchment representation

Five terrain surface features are included in the terrain image for catchment representation: elevation, slope, aspect, curvature and mask. According to Zevenbergen & Thorne (1987), the terrain surface of nine neighbouring raster cells can be approximated by a two-degree polynomial from which slope, aspect and curvature can be derived. The slope is defined as the magnitude of the gradient vector at each raster cell, representing the maximum rate of change in value from the centre cell to its neighbours and reflecting the steepness of the terrain and the overall movement rate of the water. The aspect identifies the direction of the water flow at each raster cell and is the directional component of the gradient vector. The curvature is defined as the second derivative of the polynomial where two meaningful values can be calculated: the profile curvature, which describes the acceleration and deceleration of the flow, and the plan curvature, which describes the convergence and divergence of the flow (De Smith et al., 2007). Our approach uses the difference of these two curvatures (De Smith et al., 2007) for simplicity. The mask is a binary image that indicates the catchment areas and no-data areas as 1 and -1 respectively. The above features are rescaled linearly to the range of [-1, 1] and concatenated as a multichannel terrain image. We tested our approach with only elevation and with all the features and found that introducing features makes training significantly faster.

### 4.3 CNN-based prediction model

The prediction model is designed to process data in different formats, which is usually called a joint model (Ngiam et al., 2011). As Figure 2 shows, the prediction model consists of a convolutional autoencoder (the main network) and a feedforward fully connected neural network (the subnetwork) that attaches to the latent layer of the main network. The main and subnetworks process the terrain and hyetograph data respectively. After the latent layer, the main network decodes the combined data and predicts the water depth values (metres).

The encoder of the main network is a chain of convolutional modules that consists of three convolutional layers and one pooling layer. The decoder is a chain of upsampling modules that contain 1 upsampling layer followed by two convolutional layers. The dimensions (height, width and dimensionality of features) of the input and output are $256 \times 256 \times 5$ and $256 \times 256 \times 1$ respectively. The subnetwork consists of one fully connected layer and one reshape layer. The size of the fully connected layer is 4096 for easy reshaping and concatenation to the main network. The kernel sizes are $3 \times 3$ for all the convolutional layers and $2 \times 2$ for all the pooling and upsampling layers. We use a small kernel size to preserve the thin structure of the terrain and deep layers to extend the receptive field (Luo et al., 2016). The activation functions for all the layers are Leaky-ReLU (Maas et al., 2013) to avoid the "vanishing gradient problem" (Hochreiter et al., 1998) for sigmoid units and the dead neuron problem caused by bad weight initialization for rectified linear units (Nair & Hinton, 2010).

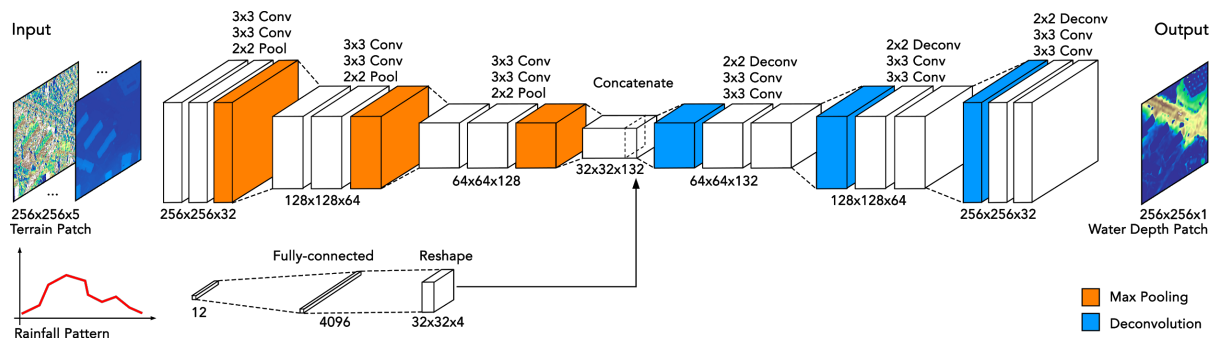

Figure 2. The flood prediction network.

Water simulation results usually contain more no-water and shallow-water areas than deep-water areas, meaning that the dataset would be imbalanced and could lower the accuracy for deep-water areas. Therefore, a weighted mean squared error instead of common mean squared error is proposed for the loss function. The definition of the loss is given as:

$$\frac{1}{n}\sum e^{y+c}(y-\hat{y})^2$$

where the weights are obtained by the exponentiation of simulated water depth $y$ plus constant $c$; $\hat{y}$ is the predicted water depth and $n$ is the number of samples. With a larger $c$, the model tends to underestimate for deep-water areas. For all the tests in this paper, we use $c = -1$.

### 4.4 Aggregating cell values from patches

As mentioned above, the final water depth prediction is aggregated from the output water depth patches whose locations (centre points) are determined by an orthogonal grid. The grid size is user-specified and should not be greater than the patch size. When the grid size equals the patch size, the boundaries of the neighbouring patches touch each other without overlaps. A smaller grid size leads to more patches that overlap with each other. Choosing the patch size is a balance between time and accuracy, as aggregating the redundant predictions from overlapping areas can reduce the impact of outliers. We present the tests of four patch aggregation options: use of mean, median, or maximum values for overlapping patches, or no patch overlaps. All results were performed with a grid size of 128 using the mean value unless mentioned otherwise.

## 5 Experimental Setup

We applied our framework to three different catchment areas located in Luzern and Zurich, Switzerland, and Coimbra, Portugal, using 1 m-resolution elevation rasters. The framework was implemented in Python using TensorFlow 1.10 (Abadi et al., 2016). All the processes, including simulation, training and validation, were performed with GPU acceleration.

### 5.1 Simulation rainfall data

To prepare the training data for the prediction model, 18 flood simulations were carried out for each catchment area using design one-hour rainfall hyetographs with rainfall return periods of 2, 5, 10, 20, 50 and 100 years (Table 1). For each return period, 1 out of 3 was randomly marked as the test set, and the rest constituted the training set. The CNN was trained using only the training set and was evaluated using the test set.

The simulations were carried out using the CADDIES cellular-automata flood model (Guidolin et al., 2016) for the Zurich and Luzern catchments and Infoworks ICM software by Innovyze, Newbury, United Kingdom (Innovyze, 2019) for the Coimbra catchment. CADDIES is a cellular-based surface flood model, whereas Infoworks ICM is a physically based model that can consider coupled pipe/surface flow in urban areas. As we already mentioned, we only stored the maximum water depth outputs in the current stage. In this work, we aimed to replicate the output of a flood simulator; therefore, we preserved the raw simulation outputs and no post-processing was conducted. Thresholds that better represent reality, for example, the use of the 90th percentile (Bruwier et al., 2018), can be made a posteriori of the prediction by the users.

Table 1. Hyetographs used for simulations

| Name | Test set | Return period | Rainfall intensity (mm/h) | | | | | | | | | | | |
|---|---|---|---|---|---|---|---|---|---|---|---|---|---|---|
| | | | 0-5 min | 5-10 min | 10-15 min | 15-20 min | 20-25 min | 25-30 min | 30-35 min | 35-40 min | 40-45 min | 45-50 min | 50-55 min | 55-60 min |
| tr2 | **yes** | 2 | 8.7 | 9.9 | 11.5 | 14.3 | 20.1 | 80.1 | 27.3 | 16.5 | 12.7 | 10.6 | 9.2 | 8.3 |
| tr5 | no | 5 | 12.3 | 13.8 | 16.1 | 19.8 | 27.6 | 104.9 | 37.2 | 22.8 | 17.7 | 14.8 | 13.0 | 11.7 |
| tr10 | **yes** | 10 | 14.9 | 16.7 | 19.4 | 23.8 | 33.0 | 120.1 | 44.1 | 27.3 | 21.3 | 17.9 | 15.7 | 14.2 |
| tr20 | no | 20 | 17.4 | 19.5 | 22.6 | 27.5 | 37.9 | 133.7 | 50.5 | 31.6 | 24.7 | 20.9 | 18.4 | 16.6 |
| tr50 | no | 50 | 20.9 | 23.3 | 26.9 | 32.6 | 44.5 | 150.4 | 58.8 | 37.2 | 29.3 | 24.9 | 22.0 | 19.9 |
| tr100 | **yes** | 100 | 24.1 | 26.8 | 30.7 | 37.0 | 50.1 | 161.4 | 65.6 | 42.1 | 33.4 | 28.6 | 25.3 | 23.0 |
| tr2-2 | no | 2 | 9.6 | 9.6 | 13.5 | 13.5 | 53.7 | 53.7 | 18.3 | 18.3 | 11.1 | 11.1 | 8.5 | 8.5 |
| tr5-2 | **yes** | 5 | 13.4 | 13.4 | 18.7 | 18.7 | 71.1 | 71.1 | 25.2 | 25.2 | 15.4 | 15.4 | 12.0 | 12.0 |
| tr10-2 | no | 10 | 16.2 | 16.2 | 22.5 | 22.5 | 82.1 | 82.1 | 30.1 | 30.1 | 18.7 | 18.7 | 14.5 | 14.5 |
| tr20-2 | no | 20 | 19.0 | 19.0 | 26.1 | 26.1 | 92.1 | 92.1 | 34.7 | 34.7 | 21.7 | 21.7 | 17.0 | 17.0 |
| tr50-2 | no | 50 | 22.7 | 22.7 | 31.0 | 31.0 | 104.6 | 104.6 | 40.9 | 40.9 | 25.9 | 25.9 | 20.4 | 20.4 |
| tr100-2 | no | 100 | 26.1 | 26.1 | 35.2 | 35.2 | 113.5 | 113.5 | 46.1 | 46.1 | 29.6 | 29.6 | 23.5 | 23.5 |
| tr2-3 | no | 2 | 8.8 | 8.8 | 8.8 | 14.5 | 14.5 | 14.5 | 42.5 | 42.5 | 42.5 | 10.7 | 10.7 | 10.7 |
| tr5-3 | no | 5 | 12.3 | 12.3 | 12.3 | 20.1 | 20.1 | 20.1 | 56.6 | 56.6 | 56.6 | 14.9 | 14.9 | 14.9 |
| tr10-3 | no | 10 | 14.9 | 14.9 | 14.9 | 24.1 | 24.1 | 24.1 | 65.7 | 65.7 | 65.7 | 18.0 | 18.0 | 18.0 |
| tr20-3 | **yes** | 20 | 17.5 | 17.5 | 17.5 | 27.9 | 27.9 | 27.9 | 74.0 | 74.0 | 74.0 | 21.0 | 21.0 | 21.0 |
| tr50-3 | **yes** | 50 | 20.9 | 20.9 | 20.9 | 33.1 | 33.1 | 33.1 | 84.6 | 84.6 | 84.6 | 25.0 | 25.0 | 25.0 |
| tr100-3 | no | 100 | 24.1 | 24.1 | 24.1 | 37.5 | 37.5 | 37.5 | 92.4 | 92.4 | 92.4 | 28.7 | 28.7 | 28.7 |

## 5.2 Training data

The training and testing data were hyetographs and patches sampled from the terrain and water depth images. As we described, for each catchment, 10,000 patch locations were randomly sampled, producing 10,000 terrain patches and 180,000 water depth patches. Water depth patches as well as hyetographs were split into training and test sets according to the information in Table 1. To study the generalizability of CNN on different hyetograph inputs, one flood prediction model was trained for each catchment area. We use identical meta parameters for training all models; specifically, we used the Adam optimizer (Kingma et al., 2014) for 200 epochs, with a batch size of 32 and a fixed learning rate of 0.0001.

## 5.3 Evaluation and validation

The performance of the proposed model was evaluated based on computational time, prediction accuracy and the ability of generalization on hyetographs. The computational times were measured by repeating the prediction process and calculating the average time. The time for necessary preprocessing (e.g., calculating the terrain features) is also reported. In addition, both the accuracy and computational time of different aggregation methods were analysed to discuss the trade-offs between speed and accuracy.

The accuracy was assessed by the mean absolute error (MAE) and the 2D histogram between all the raster cells of the predicted and simulated water depths. The MAE is defined as $\frac{1}{n}\sum_{i}^{n} |\hat{y}_i - y_i|$, where $\hat{y}_i$ and $y_i$ are the $i$-th predicted and simulated raster cells respectively. We used the MAE to assess the accuracy of the results produced by different meta parameters, such as patch aggregation methods and grid size for patch locations. The 2D histogram is a square plot with $m \times m$ pixels, where the pixel at row $i$ and column $j$ represents the number of water depth raster cells that are $y_i$ m by simulation and $y_j$ m by prediction. The water depth difference between adjacent pixels is 0.1 m. We used a 2D histogram to analyse the accuracy distribution in shallow and deep-water areas. In addition to these two assessment methods, the histogram of the error $\Delta y = \hat{y}_i - y_i$ and the spatial distribution of the relative error $\delta y_i = \Delta y_i / y_i$ are also reported.

The trained models were also validated using real rainfall events[1] that were not included in the training data to further investigate the generalizability of our flood prediction model. Real rainfall events that were less than 70 minutes were selected, clipped to 60 minutes, and resampled to 12-dimensional vectors. The accuracies were reported using histograms as well as spatial plots.

# 6 Results

Figure 3 shows the simulated and predicted water depths for the 100-year design rainfall event. We chose this hyetograph because it was the heaviest rain in our testing data and could reflect the performance of our model in extreme conditions. Unless mentioned otherwise, the results shown in this chapter correspond to this rainfall pattern.

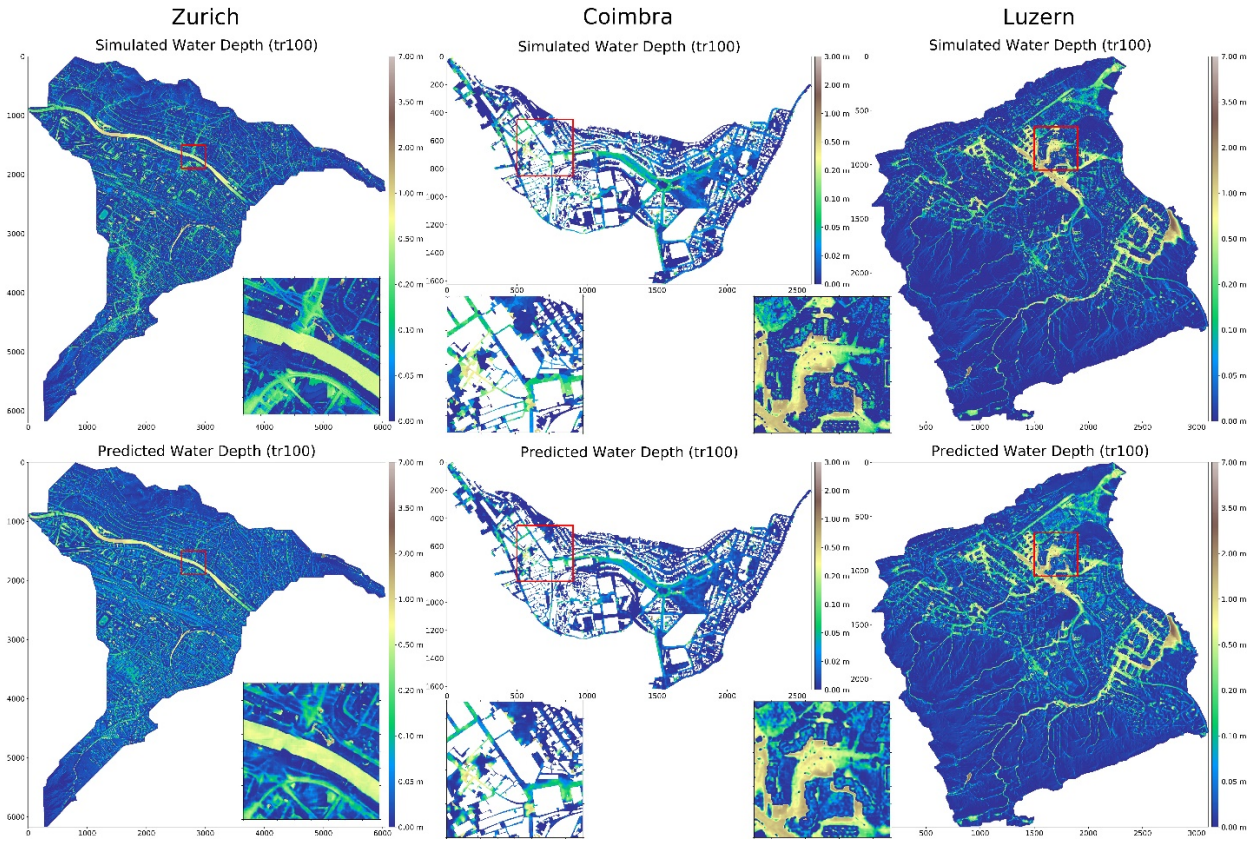

Figure 3. The simulated (top) and predicted (bottom) water depths of rainfall tr100 for the Zurich, Coimbra and Luzern catchments (left to right).

## 6.1 Prediction time

In Table 2, we present the average time of different patch aggregation methods of our approach. We found that a trained model significantly reduces the computational time for water depth prediction compared with that of the cellular automata-based models, using only 0.5% simulation time. For all three catchment areas, the no patch overlap option takes the least time.

---

[1] The rainfall events were recorded by a rain gauge owned by Aguas de Coimbra (Portugal) located within the catchment boundary.

The computation times of using the mean value and maximum value options are very close, and the time difference is less than 6.2% on average. In contrast, the median value option is the slowest due to the need to keep all the data in the memory before the median value can be obtained.

Table 2. Average time performance of the prediction model

| name | terrain image size (pixel) | terrain image pre-processing | [1] prediction time (no patch overlaps) | [1] prediction time (use mean value) | [1] prediction time (use median value) | [1] prediction time (use max value) | [1] simulation time | [2] training time |
|---|---|---|---|---|---|---|---|---|
| Luzern | 3369 × 3110 | 1.898 s | 0.678 s | 2.693 s | 14.749 s | 2.556 s | 2 h 20 min | 5 h 25 min |
| Zurich | 6175 × 6050 | 6.627 s | 1.366 s | 5.677 s | 75.12 s | 5.293 s | 4 h 54 min | |
| Coimbra | 1625 × 2603 | 0.636 s | 0.242 s | 0.965 s | 5.048 s | 0.902 s | 2 h 18 min | |

[1] The times are averaged and per rainfall event.

[2] For each catchment area, the amount of training data and training parameters were the same, and identical meta parameters were used; therefore, the average time is presented.

### 6.2 Prediction accuracy

*6.2.1 Comparison of the patch aggregation methods*

The MAEs and the error distributions of the different patch aggregation methods are presented in Figure 4, showing that the errors of different methods are all on the same level, which indicates that choosing different aggregation methods has little effect on the accuracy in general. Moreover, the results of rainfall events from the test set do not show higher MAEs than those from the training set, suggesting that our model generalizes well with rainfall variations. A more detailed level shows that using overlapping patches in general results in lower MAEs than those with the no overlaps option. Using the median value usually gives better results than the mean and maximum options. In addition, the histograms of the prediction error on the right show that the no overlap option makes more underpredictions and overpredictions in all catchments, suggesting more outliers.

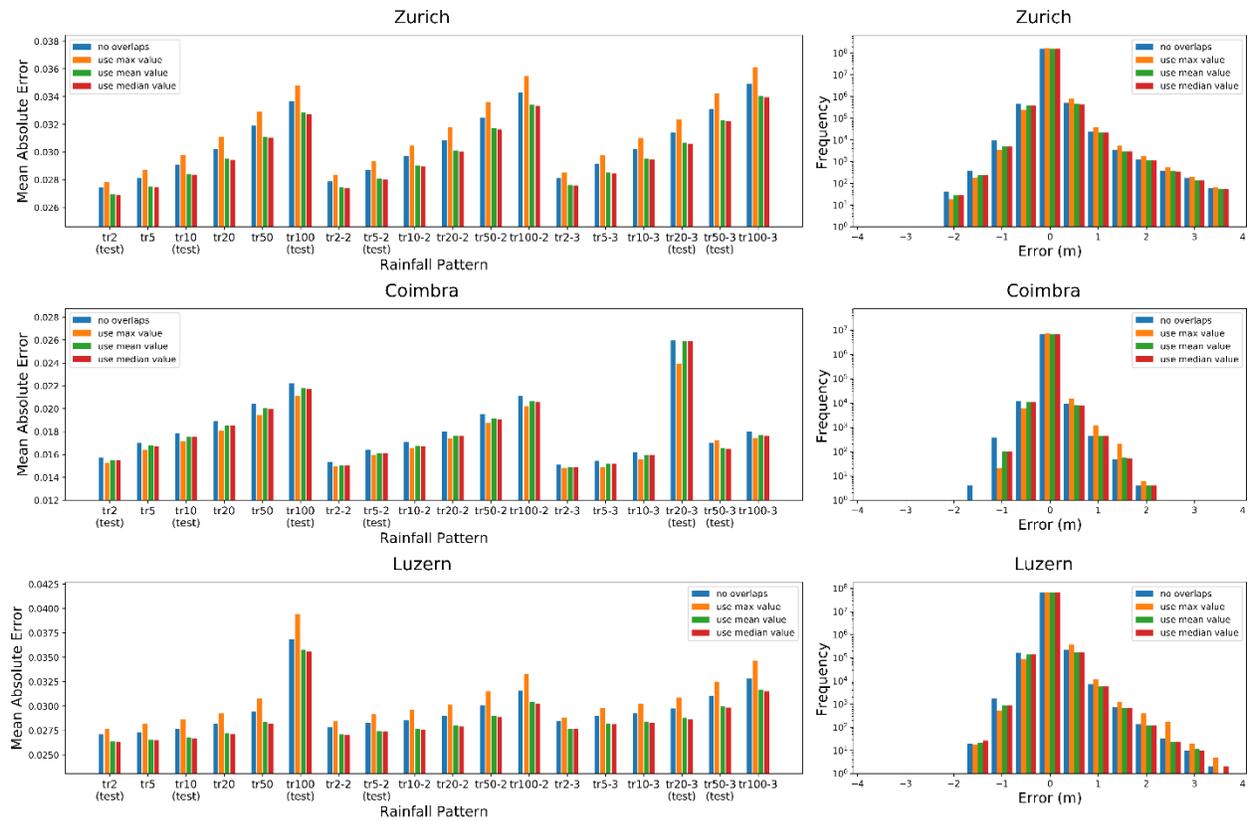

Figure 4. The mean absolute errors of each hyetograph (left) and the error histograms of all hyetographs (right) in the Zurich, Coimbra and Luzern catchments (top to bottom) using different aggregating methods.

*6.2.2 Accuracy in shallow and deep waters*

The prediction accuracy of our model is presented as 2D histograms in Figure 5, showing the density of the water depth predictions and simulations ranging from 0.0 m to 7.0 m in 0.1 m resolution. The plot pixel in row $i$ and column $j$ represents the number of water depth raster cells that are $y_i$ m by simulation and $y_j$ m by prediction; therefore, a model with 100% accuracy would produce a plot in which all pixels except the diagonal are 0 (dark). The more divergent the plot is from its diagonal, the lower the prediction accuracy. From the figure, it is clear that all coloured pixels match the diagonal well, indicating that our model makes accurate predictions in both shallow-water and deep-water areas. Note that the plots are coloured logarithmically to show small numbers in the shallow-water parts. The imbalanced distributions of cells with different water depths are also reported in this figure.

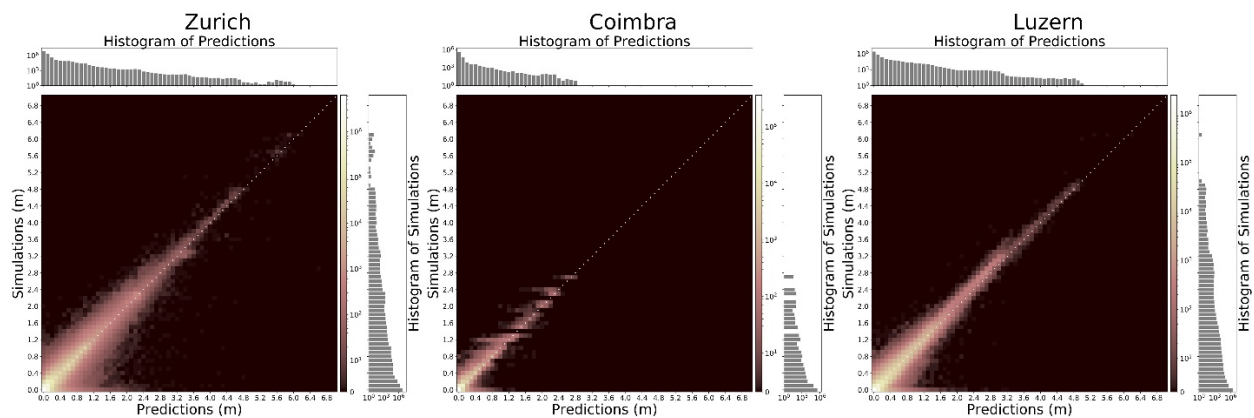

Figure 5. 2D histogram for the water depth prediction (tr100) rounded by 0.1 m in the Zurich, Coimbra and Luzern catchments (left to right). Note that the plots are coloured logarithmically.

We noticed that the first rows in Figure 5 indicate that some raster cells that are 0 m water depth by simulation were overpredicted by maximum 2 to 3 m errors. To validate the accuracy of our model on these raster cells, we zoomed in to where the errors are higher than 1 m. An example of these enlargements is presented in Figure 6, which clearly shows that the high-error cells are due to the "smoothing" our model tried to make around the sharp water depth changes by simulations. These sharp changes may be caused by artefacts in the elevation data. Considering the size of the entire catchment and the accuracy prediction of the deep-water cell at the centre of the enlargement, these errors do not seem to affect the accuracy of our model or the interpretation of the prediction results. A detailed counting of the high-error cells and areas (group of cells with distance between each other less than 16 pixels) is shown in Table 3. Note that these numbers are considered small as each catchment contains millions of raster cells.

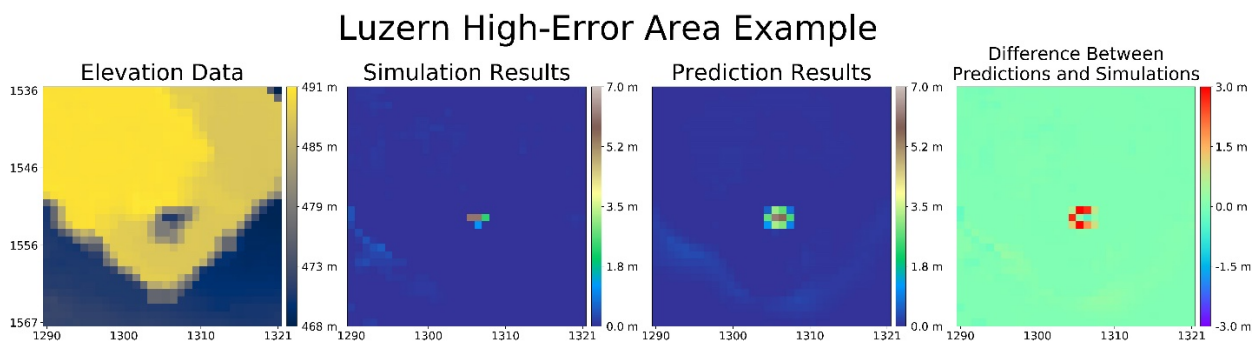

Figure 6. The enlargement of the area with the highest prediction error in the Luzern case.

Table 3. Counting results of high-error cells of rain tr100 in all catchments

| Absolute errors (m) | Zurich | | Coimbra | | Luzern | |
| --- | --- | --- | --- | --- | --- | --- |
| | Number of cells | Number of areas | Number of cells | Number of areas | Number of cells | Number of areas |
| [0.5, 1) | 7949 | 4160 | 210 | 59 | 3140 | 1600 |
| [1, 2) | 835 | 354 | 5 | 4 | 212 | 101 |
| [2, 3) | 57 | 28 | 0 | 0 | 10 | 4 |
| [3, ∞) | 10 | 4 | 0 | 0 | 2 | 1 |

*6.2.3 Spatial distribution of errors*

Figure 7 shows the spatial distribution of errors and relative errors of the three catchment areas. As seen, there are no significant differences between shallow-water and deep-water areas, indicating that the performance of the model is not biased relative to the water depth prediction.

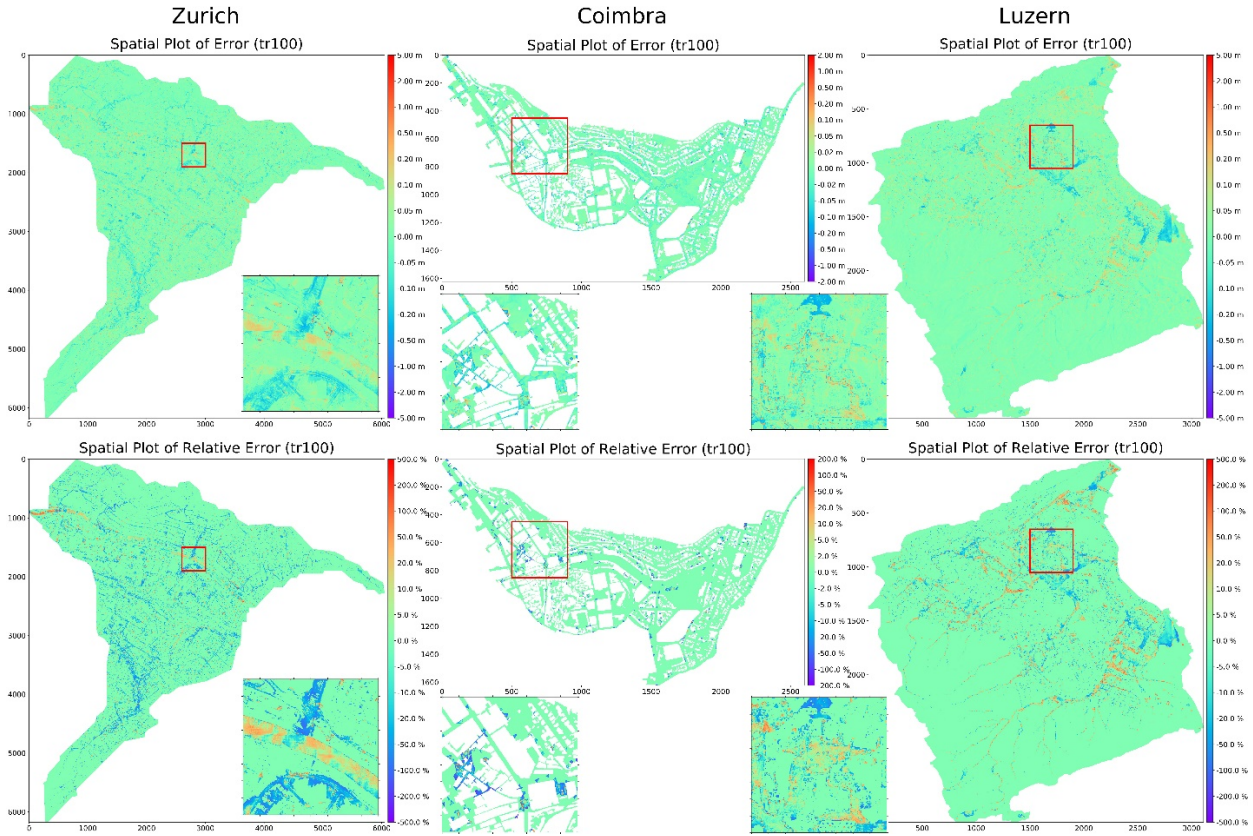

Figure 7. Errors (top) and relative errors (bottom) of rainfall tr100 for all catchment areas.

## 6.3 Validation with real rains

The generalization of CNN-based floods on hyetographs makes it useful for predicting floods from arbitrary rainfall patterns. As shown in Figure 8, three predictions were made and compared with simulation results using real rainfall data in Coimbra. The error histograms show that the model successfully handled real rainfall events with one or multiple peak values and made accurate predictions.

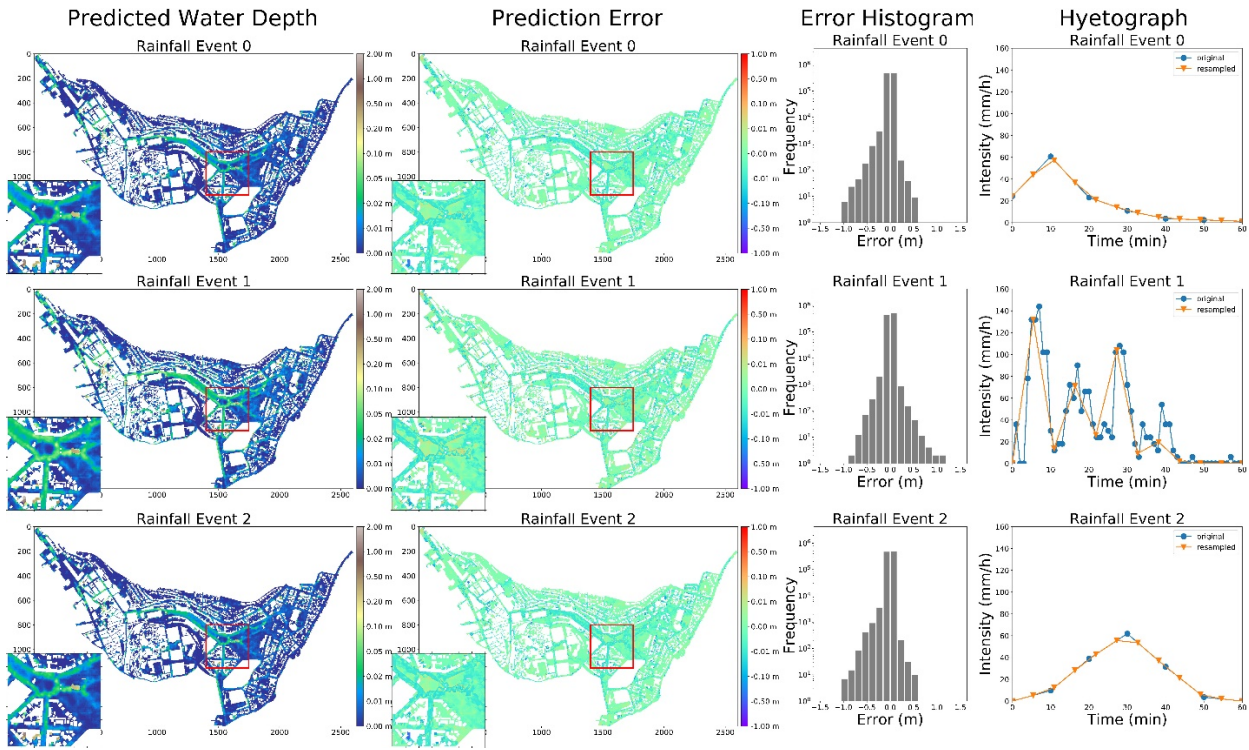

Figure 8. The prediction results of three real rainfall events for the Coimbra case.

## 7 Conclusions and Future Steps

Computational complexity has been the bottleneck of performing systematic flood analyses with physically based models on large-scale, high-resolution scenarios for flood risk management. Regarding this challenge, this paper has proposed that, rather than conducting physically based simulations, the maximum water depth can be generated by CNNs using the information learned from data. In this study, this approach achieved a major increase in speed and maintained promising accuracy. The approach was tested in three different catchment areas, and the results showed that the improvement in computational time was significant. A trained model takes only 0.5% of the time compared with that of cellular automata-based models. Additionally, unlike physically based models, which could potentially take 10 times more computational time when doubling the raster resolutions (Guidolin et al., 2016), the time increase of CNN models is expected to be linear as it only correlates with the number of patches. This could be another driver for further investigating the potential of CNN models.

In addition to the increased speed, investigations of different combinations of meta parameters and configurations showed that the CNN model achieved high accuracy in both shallow- and deep-water areas even when the data were imbalanced. This is supported by the tests made with real rainfall events that did not exist in the training data, suggesting that the CNN model generalized well regarding input rainfall variations.

For future works, it seems very interesting and promising to use different combinations of the inputs and outputs of the CNN model, for example, using different input terrains while keeping the hyetograph unchanged or introducing other features such as water velocity. These can be the immediate next steps for this research. Another interesting direction would be to train and validate the model using observational data instead of simulations, which is now becoming possible through the combination of crowdsourcing methods (Zheng et al., 2018) and computer vision techniques (Moy de Vitry et al., 2019).

## Acknowledgments, Samples, and Data

This work was funded by the China Scholarship Council grant 201706090254, and the authors would like to thank Água de Coimbra for providing the drainage network information used in this study. The simulation data, source code and trained models can be obtained from the data repository (Guo et al., 2019) hosted by the research collection of ETH Zurich with DOI link 10.3929/ethz-b-000365484. The authors declare no conflicts of interests.